# Combining Symbolic and Numeric Approaches to Uncertainty Management

Bruce D. D'Ambrosio
Department of Computer Science
Oregon State University

June 4, 1987

**Abstract**

Uncertainty is represented in an Assumption-based Truth Maintenance System (ATMS) by tokens called *assumptions*, which are used to represent belief in uncertain facts. Given a set of assumptions and a set of inferences which can be drawn from these assumptions and their consequents, the ATMS derives a complete boolean expression (label) for the truth value of every proposition in the database, expressed in terms of the original assumption tokens. Thus, an ATMS can be viewed as a symbolic algebra system for uncertainty reasoning. Previously, assumptions have always been taken to be truth variables ranging over boolean truth values. This paper describes a method of attaching numeric certainty estimates to assumptions, and deriving numeric truth values from the labels of ATMS propositions. This technique has several major advantages over conventional methods for performing inference with numeric certainty estimates, including improved management of dependent and partially independent evidence, faster run-time evaluation of propositional certainties, and the ability to query the certainty value of a proposition from multiple perspectives.

## 1 Background

Current techniques for performing inference with numeric certainty values in expert systems [Buchanan and Shortliffe, 84], [Lowrance, 86], rely on numeric combination of evidence at each stage of inference. All known evidence is reduced to a single number or pair of numbers associated with each proposition. This means that the source of each derived support is lost. This, combined with a lack of information regarding the necessary conditional probabilities, requires that independence assumptions be made in order to combine evidence from multiple inferences supporting a single proposition [Cheeseman, 85]. Further, it is not possible to determine the sensitivity of consequent certainty values to individual pieces of evidence without re-doing the entire inference process. Finally, the simple numeric schemes typically used in expert systems usually do not allow for changes in support for antecedents once inference has taken place.

An alternative paradigm for reasoning under uncertainty [Cohen, 84], [deKleer, 86], [Doyle, 79], [McAllester, 80] can be traced back to the work of Doyle [Doyle, 79] on symbolic truth maintenance systems. One of the most recent of these systems, the Assumption-based Truth Maintenance System (ATMS) of deKleer [deKleer, 86] builds symbolic expressions (*labels*) for the truth value of all propositions in a database. Specifically, the label of a proposition is a list of sets of assumptions, where the proposition is held to be true if all of the assumptions in at least one of the sets are true. While this approach automatically eliminates dependent evidence and provides for sensitivity analysis of resulting truth values, it suffers from two limitations. First, the final propositional truth values are essentially drawn from only three possible values - false, true, and uncertain. This is inadequate for applications in which it is necessary to rank alternatives. Second, the computational complexity of maintaining the ATMS database grows rapidly with the size of the database and the interconnectedness of the propositions.

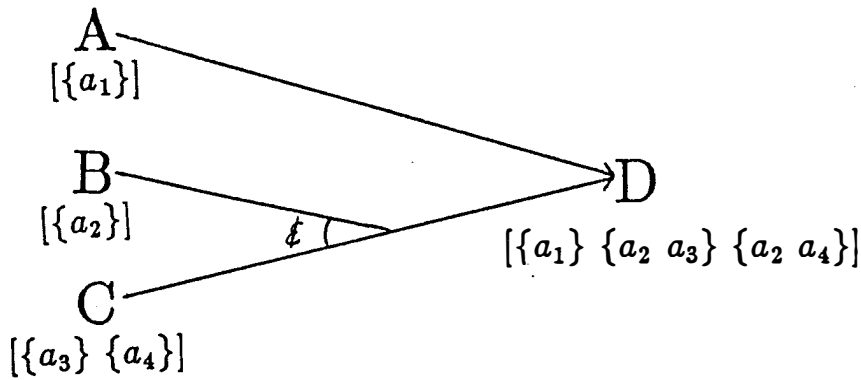

Figure 1: Simple Label Propagation in an ATMS

## 2 Overview

We are engaged in research on a novel method for computing the certainty of derived propositions from numeric certainty information for the initial evidence, based on combining an ATMS with the evidential combination algorithms of the Dempster/Shafer theory of evidence [Shafer, 76]. The method relys on the propagation mechanisms in the ATMS to perform most evidential inference and combination operations symbolically, and only substitutes numeric values when asked for the certainty of a proposition. Advantages include improved handling of non-independent evidence, very fast run-time evaluation of propositional certainties, and the ability to request the certainty value of any proposition from several perspectives. Also, changes in support for antecedents are automatically reflected in consequent propositions, even after inference has occurred.

One view we can take of an ATMS is as a symbolic algebra system for uncertainty information. That is, the ATMS starts with a set of uncertain data and a set of inferences which can be drawn from that data. It then computes a closed form symbolic expression for the truth of all consequent propositions, in terms of the symbolic truth values (*assumptions*) of the original data. One consequence of this capability is that, once problem solving is complete, the truth of a proposition in any solution or partial solution can readily be determined with inexpensive, purely local operations.

As mentioned earlier, the ATMS has only limited ability to rank alternative hypotheses. If we view ATMS assumptions as certainty variables, and can establish a mapping between the computations performed by the ATMS and those of some well-founded numeric certainty calculus, then we can substantially improve its ability to rank alternatives. We are establishing such a mapping for the Dempster/Shafer theory of evidence [Shafer, 76] and a simplification/extension of it, Support Logic Programming [Baldwin, 85]. The elements which must be mapped are basic probability assignments, primitive hypothesis sets, frames of discernment, inference rules (SLP) (and associated algorithms for propagating evidence across inferences), and evidence combination algorithms. In the remainder of this paper we briefly review Assumption-based Truth Maintenance and Support Logic Programming, detail the mapping we have established, and illustrate operation of the combined reasoner on a simple example. We conclude with a description of our continuing research.

## 3 Details

### 3.1 Assumption-Based Truth Maintenance

An ATMS is a form of propositional truth maintenance system in which all of the assumptions supporting a proposition are explicitly recorded with it. Specifically, the *label* of a proposition is a list of sets of assumptions, where the proposition is held to be true if all of the assumptions in at least one of the sets are true.

A proposition gets its label via assumption-set propagation through propositional instantiations of inferences called *justifications*, as illustrated in fig. 1. This propagation can be considered a two stage process. First, each new assumption set is propagated through each justification for which the proposition it supports is an antecedent. Second, the assumptions sets arriving at consequent propositions are combined with those already in the label set for the consequent.

Each assumption set generated by a justification is compared with the assumptions sets already in the label of the consequent proposition. If the new assumption set is identical with or a superset of any assumption set already in the label, it is discarded. If it is a subset of any assumption set in the label, its superset assumption sets are removed from the label. Finally, if the new assumption set was not discarded, then it is added to the label of the consequent proposition and passed on as input to all justifications which have that proposition as an antecedent, starting a new cycle of propagation. Various heuristics are necessary to ensure that assumption set propagation is efficient.

The conclusion of a justification can be the special proposition false, and all assumption sets which get propagated to false are marked as inconsistent. When an assumption set is marked inconsistent, it and all its supersets are removed from any labels in which they appear. Also, labels generated by conjunctive justifications (justifications with more than one antecedent) are checked for consistency before propagation to consequent propositions, and discarded if inconsistent. This checking is performed by maintaining a database of canonical forms of assumption sets with consistency information.

This inconsistency management is used, for example, to ensure that mutually inconsistent propositions are never combined to derive support for any other proposition. Figure 2 shows an example of how a mutually exclusive set of propositions can be represented in an ATMS using assumptions and justifications to false.

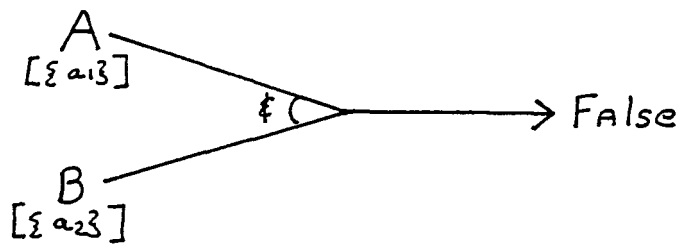

Figure 2: Representing mutual exclusion in an ATMS

This technique can be used to partially represent *one-of disjunctions*. By a one-of disjunction we mean a mutually exclusive and exhaustive set of propositions, or roughly the equivalent of a Dempster/Shafer *primitive hypothesis set*. The propagation and inconsistent label management algorithms can be shown to implement a nearly complete propositional logic. While the representation in Fig. 2 does imply that no more than one of the propositions can be true, however, it does not imply that exactly one must be true. This aspect of one-ofs is handled separately, by explicitly representing disjunctions in a separate data structure. The process of finding consistent assignments of truth values to assumptions is called interpretation construction. Each such assignment is called an interpretation, and must assign true to exactly one assumption from each disjunction, consistent with all justifications installed.

## 3.2 Support Logic Programming

The ATMS is an elegant and efficient way to simultaneously explore all solutions within a search space. However, once the exploration is complete, it can only provide three valued estimates of the truth of derived propositions: false, true, or uncertain. Thus, it provides little support for ranking alternatives for decision making. Support Logic Programming (SLP), a system recently introduced by Baldwin and based on the Dempster/Shafer theory of evidence, offers the potential of increasing the amount of information available from an ATMS label by providing numeric certainties for derived propositions. While SLP does not support the complete Dempster/Shafer model of evidential reasoning, it will provide an adequate starting point for our discussion. Later, we will discuss how this can be extended to include the full Dempster/Shafer model.

Support logic programming is a prolog-like programming system in which the uncertainty associated with facts and rules is represented by a pair of supports. For example:

    (active reaction): [1/4, 2/3]

is interpreted to mean that the *support* for the proposition (active reaction) is 1/4, and the support for (not (active reaction)) is 1/3. While the support for a proposition and its negation cannot add to greater than 1, they may add to less than one. The difference between the sum and one represents the degree to which we are uncertain about the likelihood of the proposition. There are two statement types[1]:

```
(1) P : [Sl(P), Su(P)]
 e.g.: (active reaction): [1/4, 2/3]

(2) P :- Q : [Sl(P|Q), Su(P|Q)]
 (dir reactants down):- (active reaction): [1, 1]
```

The first states that proposition P is a fact with the specified support-pair, and the second states an inference rule and associates a support pair describing the belief in the validity of the inference. Inference antecedents can be composite, and inference is done as follows:

```
P :- Q : [Sl(P|Q), Su(P|Q)]
Q   : [Sl(Q), Su(Q)]}
P   : [Sl(P|Q).Sl(Q),1-(1-Su(P|Q)).Sl(Q)]

 (dir reactants down) :- (active reaction): [2/3, 1]
 (active reaction): [1/4, 2/3]
 (dir reactants down): [1/6, 1]
```

Finally, supports can be combined:

```
P : [S1, U1]
P : [S2, U2]
P : [S, U]

where S = (S1*U2 + S2*U1 - S1*S2) / K,
      U = 1 - ((1-U1)(1-S2) + (1-U2)(U1-S1)) / K
and   K = 1 - S2*(1-U2) - S1*(1-U2)

 (active reaction) : [1/4, 2/3]
 (active reaction) : [1/2, 3/4]
 (active reaction) : [22/37, 34/37]
```

Support Logic Programming provides a representation for numeric certainties and algorithms for propagating those certainties through inferences to derive support for consequents. Additionally, it provides a method for combining multiple supports for a single proposition into a single composite support. However, the Dempster/Shafer theory provides for evidence over wider sets of alternatives than a proposition and its negation. Specifically, evidence can be given for members of an arbitrary *frame of discernment*, the power set of a primitive hypothesis set. Prolog has no direct means of representing this wider evidential frame. In the next section we show how a larger subset of the Dempster/Shafer theory can be mapped into the ATMS.

### 3.3 Mapping SLP into the ATMS

We use an ATMS *assumption* to carry each element of a basic probability assignment. In the Dempster/Shafer theory of evidence, a source provides evidence not for a single proposition, but rather distributes evidential mass over an entire *frame of discernment*. Our current implementation allows only evidence for primitive hypotheses and their negations. We use the ATMS One-of disjunction to represent a primitive hypothesis set, and the assumption supporting each hypothesis is tagged with the evidential mass assigned to the hypothesis (separate assumptions are created for negation

---

[1]Notation in SLP follows Prolog standards, :- is read "is implied by", X,Y should be read "X AND Y", and X;Y should be read "X OR Y"

hypotheses, again tagged with the evidential mass assigned). Dempster/Shafer theory does not include a model for inference rules; however, SLP does. The SLP inference rule can be mapped into two ATMS justifications. Given:

```
B :- A : [Sl(B|A), Su(B|A)],
```

then we install two ATMS justifications, one from A to B supported by $Sl(B|A)$, and one from A to (not B) supported by $(1 - Su(B|A))$, as shown in fig. 3.3.

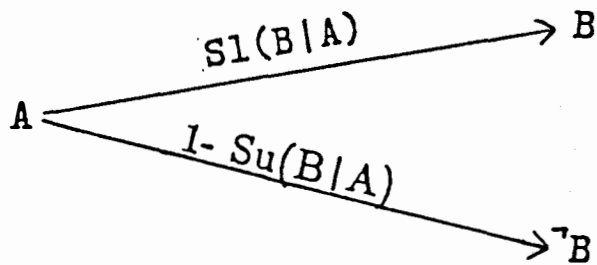

Figure 3: Representing an SLP inference rule in the ATMS

An important feature of this mapping is that the composition of a sequence of two justifications is parenthesis free[2]. As a result, we can defer numeric evaluation, use the standard ATMS assumption propagation algorithm, and compute the support pair for any proposition by examining the labels of all propositions in its frame of discernment.

Dempster/Shafer theory provides a mechanism for combining sets of evidence over a frame of discernment consisting of a primitive set of exhaustive and mutually exclusive hypotheses. Unfortunately, the basic theory requires enumerating all subsets of the power set of the primitive hypothesis set, a computationally prohibitive technique for large primitive hypothesis sets. We can use the inferential capability of the ATMS to reduce the frame of discernment to a set of simple two element frames, one for each of the primitive hypotheses and its negation, and combine evidence over each simpler frame numerically, as described below. This is analogous to Barnett's method [Barnett, 81], in that we discard (assign to theta) evidence for elements of the power set of the hypothesis space which are neither primitive hypotheses nor their negations. The difference is that we can use the ATMS label propagation algorithms to perform most of the algorithm symbolically.

We then compute the support for each proposition by numerically combining the support for the assumptions in its label and the label of its negation. Each assumption carries only a single number, equivalent to the lower support value. We extend this by providing a one as the upper support. Assumptions within a single assumption set are combined using the AND combination rule of SLP (SLP actually provides several models for AND, we use the multiplication model). This yields a numeric value for the evidence contributed by each assumption set. SLP combines these using the Dempster/Shafer combination rule. Since we have the actual assumptions underlying each piece of evidence (assumption set), we can combine them more intelligently than purely numeric approaches. Specifically, dependent evidence is automatically subsumed by the ATMS label propagation algorithm, mutually exclusive evidence can be detected by noting whether or not two assumption sets contain different members of a one-of disjunction, and the remaining assumption sets can be assumed independent and combined using the SLP combination algorithm. We are also examining the possibility of developing heuristics can be developed to examine the structure of partially independent assumption sets and make more appropriate combination decisions. In general, such a heuristic has as input data a proposition label, the numeric support value for each assumption in the label, and the declarations of mutual exclusivity among assumptions. The output of such a heuristic would be a plan for reducing the label to a support pair by pairwise combination of the assumption sets, with a specific choice of the combination rule to be used at each step.

---

[2]Note that $(1 - Su(B|A)) = Sl(Not(B|A))$. All upper supports are represented as lower support for the negation. That is why we claim the mapping is parenthesis free.

This technique only represents primitive hypotheses and their negations. We are also developing symbolic techniques for representing the full power set of the primitive hypothesis set, as well as hierarchical subsets of the power set.

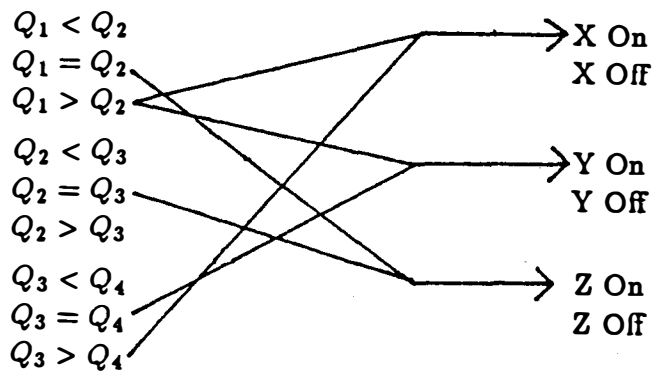

Figure 4: Graphical representation of ATMS example

Notice that we can now perform a sensitivity analysis on propositional support. This can be done by selecting assumption sets to be included in the support computation according to some specification, or by substituting extreme values for the numeric support for a particular assumption prior to evaluation. One method of selecting assumption sets is to specify a kernel set of assumptions. An assumption set can then be included in the support computation according to various measures of its consistency with the kernel set.

### 3.4 Example

We now present an example of use of the extended ATMS. The example is a simple assessment task in which we have a system consisting of three components and four numeric parameters. Each of the components can be in one of two states, ON or OFF. Each component state is determined by the relative magnitudes of certain of the numeric parameters, as shown in Fig. 4. For example, we know that if (Q1 > Q2) and (Q3 > Q4), then component X must be ON. The results shown were obtained from an early prototype we implemented for initial explorations of the ideas presented here.

The problem is to determine possible overall system states given some uncertain evidence regarding parameter magnitude orderings. We have converted the ATMS code into a Prolog-like syntax for consistency with the SLP discussion. Fig. 4 shows the inferential structure we are constructing. Fig. 5 shows a simplified version of the code, and Fig. 6 shows the output generated when it is run.

## 4 Further Research

**Completing the Dempster/Shafer Model** We are currently developing mappings for the full hypothesis space, as well as hierarchical subsets of it. Also, our current model only allows for evidence over the frame consisting of a single hypothesis and its negation. We are extending this to allow for evidence over wider frames of discernment as well.

**Other Numeric Uncertainty Models** We have shown that deKleer's ATMS can be used as a symbolic algebra system for uncertainty calculations consistent with at least one version of an uncertainty calculus based on the Dempster/Shafer theory of evidence. We have not established, though, how wide a class of numeric certainty calculation algorithms can be captured using this approach. One criteria we did identify is that the representation used in the ATMS must be such that the calculations are associative, since ordering and nesting information is lost in assumption-set propagation. We believe that of probabilistic approach, a fuzzy logic [Zadeh, 79], and the MYCIN certainty factor method, can all be adapted to representation symbolically in the ATMS.

```
;; one-of disjunctions for each component
(one-of (x on) (x off))
(one-of (y on) (y off))
(one-of (z on) (z off))

;; quantity orderings and subsystem states
(Q1 > Q2) & (Q3 > Q4)   <->  (x on)
(Q1 > Q2) & (Q3 = Q4)   <->  (y on)
(Q1 = Q2) & (Q2 = Q3)   <->  (z on)

(one-of (Q1 < Q2) (Q1 = Q2) (Q1 > Q2))
(one-of (Q2 < Q3) (Q2 = Q3) (Q2 > Q3))
(one-of (Q3 < Q4) (Q3 = Q4) (Q3 > Q4))

;; evidence about orderings
(Q1 < Q2): [0, 0]  (Q1 = Q2): [.5, .8]  (Q1 > Q2): [.3, .6]
(Q2 < Q3): [0, 0]  (Q2 = Q3): [.1, .6]  (Q2 > Q3): [.8, 1]
(Q3 < Q4): [0, 0]  (Q3 = Q4): [.6, .9]  (Q3 > Q4): [.2, .5]
```

Figure 5: Example Problem Setup

**Viewing Numeric Certainties as Summaries** The above research is predicated on a view in which numeric certainty values are taken as primitive, and ATMS assumptions are merely truth variables. This view permits us to rectify a fundamental limitation of the ATMS, the inability to rank alternatives. However, there is another problem with the pure symbolic approach to uncertainty management embodied in the ATMS, namely computational complexity. The ATMS algorithms compute the consequences of any change (addition of a proposition, justification, or assumption) for every proposition in the database. Worst-case analysis indicates this is at best linear in the number of propositions (p), and in most implementations is also no better than linear in the number of assumptions (a). Since both increase with increasing problem size, the overall algorithm is at best $O(p * a)$.

The fundamental cause of this problem is the same feature of the ATMS that provides its richness and power, namely that certainty in a proposition is represented by a complex, structured set of tokens. One way to solve this problem is to *summarize* the certainty in a more compact form, and relax the commitment to immediately compute all consequences of a change in support for every proposition. Many problems can be partitioned into a series of steps, where each step can be viewed as a cycle of alternative development followed by selection.

We plan to address the complexity problem by partitioning the ATMS into separate databases for each reasoning task. Inferences in one partition dependent on hypotheses located in another partition will be linked by creating an *assumption* in the dependent partition. The assumption will be tagged with the numeric support value obtained from the hypothesis in the antecedent partition. A forward link from the hypothesis to the new assumption can also be kept, so that the assumption can be informed of *significant* changes in support for the antecedent proposition. This propagation between partitions, however, is not done automatically by the ATMS, but rather by the higher-level problem-solver.

```
Solution: (Z OFF) (Y OFF) (X ON):   [.59, .77]
Solution: (Z OFF) (Y ON)  (X OFF):  [.01, .12]
Solution: (Z ON)  (Y OFF) (X OFF):  [.13, .29]
Solution: (Z OFF) (Y OFF) (X OFF):  [.02, .13]
```

Figure 6: ATMS Example Results

## 4.1 Search Control

The availability of numeric evaluations of alternatives significantly impacts the architecture of problem-solvers built using an ATMS. As discussed earlier, and ATMS is an elegant mechanism for exploring a set of alternatives, but provides limited mechanisms for ranking them. In part for this reason, previous work using an ATMS has stressed *completeness*. Alternatives are explored whenever there is any evidence that an alternative is possible (i.e., when the label for the corresponding proposition is non-null). This is the basic strategy available in the prototype rule-based problem-solver we have implemented, as well as the one used in the commercially available tool, ART. With numeric estimates, at least two alternate heuristic search strategies suggest themselves[3]:

1. Pursue the currently highest ranked alternative(s).

2. Pursue all alternatives with rankings above some (perhaps dynamically adjustable) threshold.

## 5 Summary

Uncertainty is a pervasive problem underlying much of AI - we rarely have complete information, and even when we do, can only afford to apply it in its entirety in toy problems. Once information has been gathered and inference is complete, decisions must be made. Numeric methods for managing uncertainty provide a means of ranking alternatives, but do not directly support alternative development. Symbolic methods directly support alternative development, but provide little guidance in selection. By combining these two classes of techniques, we are developing a system that directly supports both aspects of reasoning under uncertainty, and does so in an efficient and effective fashion.